\documentclass{article}
\usepackage[preprint]{neurips_2024}
\usepackage{graphicx}
\usepackage[utf8]{inputenc} % allow utf-8 input
\usepackage[T1]{fontenc}    % use 8-bit T1 fonts
\usepackage{hyperref}       % hyperlinks
\usepackage{url}            % simple URL typesetting
\usepackage{booktabs}       % professional-quality tables
\usepackage{amsfonts}       % blackboard math symbols
\usepackage{nicefrac}       % compact symbols for 1/2, etc.
\usepackage{microtype}      % microtypography
\usepackage{xcolor}         % colors

\usepackage{wrapfig}
\usepackage{tabularx}

\title{Depth-Wise Activation Steering for Honest Language Models}

\author{
  Gracjan~Góral\thanks{Equal contribution.
  Email correspondence to gp.goral@uw.edu.pl.
  Work done as part of MARS (Mentorship for Alignment Research Students).}\\
  University of Warsaw \\
  MARS\\
  \And
  Marysia~Winkels\footnotemark[1]\\
  MARS \\
  \And
  Steven~Basart\\
  Center for AI Safety\\
}

\begin{document}

\maketitle

\begin{abstract}
Large language models sometimes assert falsehoods despite internally representing the correct answer—failures of honesty rather than accuracy—which undermines auditability and safety. Existing approaches largely optimize factual correctness or depend on retraining and brittle single-layer edits, offering limited leverage over truthful reporting. We present a training-free activation steering method that weights steering strength across network depth using a Gaussian schedule. On the MASK benchmark—which separates honesty from knowledge—we evaluate seven models spanning the LLaMA, Qwen, and Mistral families and find that Gaussian scheduling improves honesty over no-steering and single-layer baselines in six of seven models. Equal-budget ablations on LLaMA-3.1-8B-Instruct and Qwen-2.5-7B-Instruct show the Gaussian schedule outperforms random, uniform, and box-filter depth allocations, indicating that how intervention is distributed across depth materially affects outcomes beyond total strength. The method is simple, model-agnostic, requires no finetuning, and provides a low-cost control knob for eliciting truthful reporting from models’ existing capabilities.\footnote{See \url{https://github.com/marysia/gaussian-activation-steering}. for code and experiments.}
\end{abstract}
\section{Introduction}
Large language models can produce statements that contradict what they earlier implied or internally represented to be correct. When this occurs, the failure is not a deficit of world knowledge but a breakdown in truthful reporting—\emph{honesty}. This distinction matters for auditability and safety: models that \textit{know but misreport} can evade oversight, facilitate manipulation, and degrade trust even when their factual knowledge is strong \citep{zou2023universaltransferableadversarialattacks,shen2024donowcharacterizingevaluating}. Recent work shows that even aligned systems can be pushed into unsafe or deceptive behavior by transferable prompts and in-the-wild strategies, underscoring the need for controls that directly target truthful reporting rather than only factual accuracy.

Standard countermeasures concentrate on three levers. Training-time alignment methods—such as reinforcement learning from human feedback (RLHF) \citep{ouyang2022traininglanguagemodelsfollow} and Constitutional AI \citep{bai2022constitutionalaiharmlessnessai}—can improve helpfulness and reduce overt harms, but they require backward passes, curated data, and nontrivial compute budgets, and they entangle honesty with distribution-specific preferences learned during fine-tuning. External safety classifiers provide a separate moderation layer \citep{inan2023llamaguardllmbasedinputoutput} but add engineering complexity and can be bypassed when models are used without the wrapper. Prompt-based instruction templates \citep{ZhengY0M0CHP24} are cheap to deploy but brittle against adaptive adversaries. None of these interventions provides a simple, test-time knob that directly and robustly steers a model toward truthful self-reporting.

Representation engineering offers such a knob. In representation engineering, or \emph{activation steering} as it is also known, one intervenes during inference by adding a vector to the residual stream to push generation toward or away from a target property (e.g., sycophancy, toxicity, hallucination, or refusal) \citep{turner2024steeringlanguagemodelsactivation,rimsky2024caa}. Despite impressive case studies, most depth-wise implementations are \emph{degenerate}: either a point-mass edit at a single chosen layer \citep{turner2024steeringlanguagemodelsactivation,rimsky2024caa}, or (when intervening broadly) a near-uniform weighting over many layers \citep{zhao2024gcs}. Methods that adapt coefficients—learned activation scalars \citep{stoehr2024activationscaling} and semantics-adaptive dynamic directions \citep{wang2024sadi}—optimize per-location weights but stop short of prescribing an analytic, interpretable \emph{schedule over depth}. Feature-level approaches based on sparse autoencoders operate in a different basis, improving edit specificity but not modeling how intervention strength should vary across layers \citep{obrien2024saerefusal,chalnev2024saets}. As a result, the \emph{distribution of steering strength across depth} remains an underexplored design degree of freedom.

\begin{figure}[t!]
    \centering
    \includegraphics[width=\linewidth]{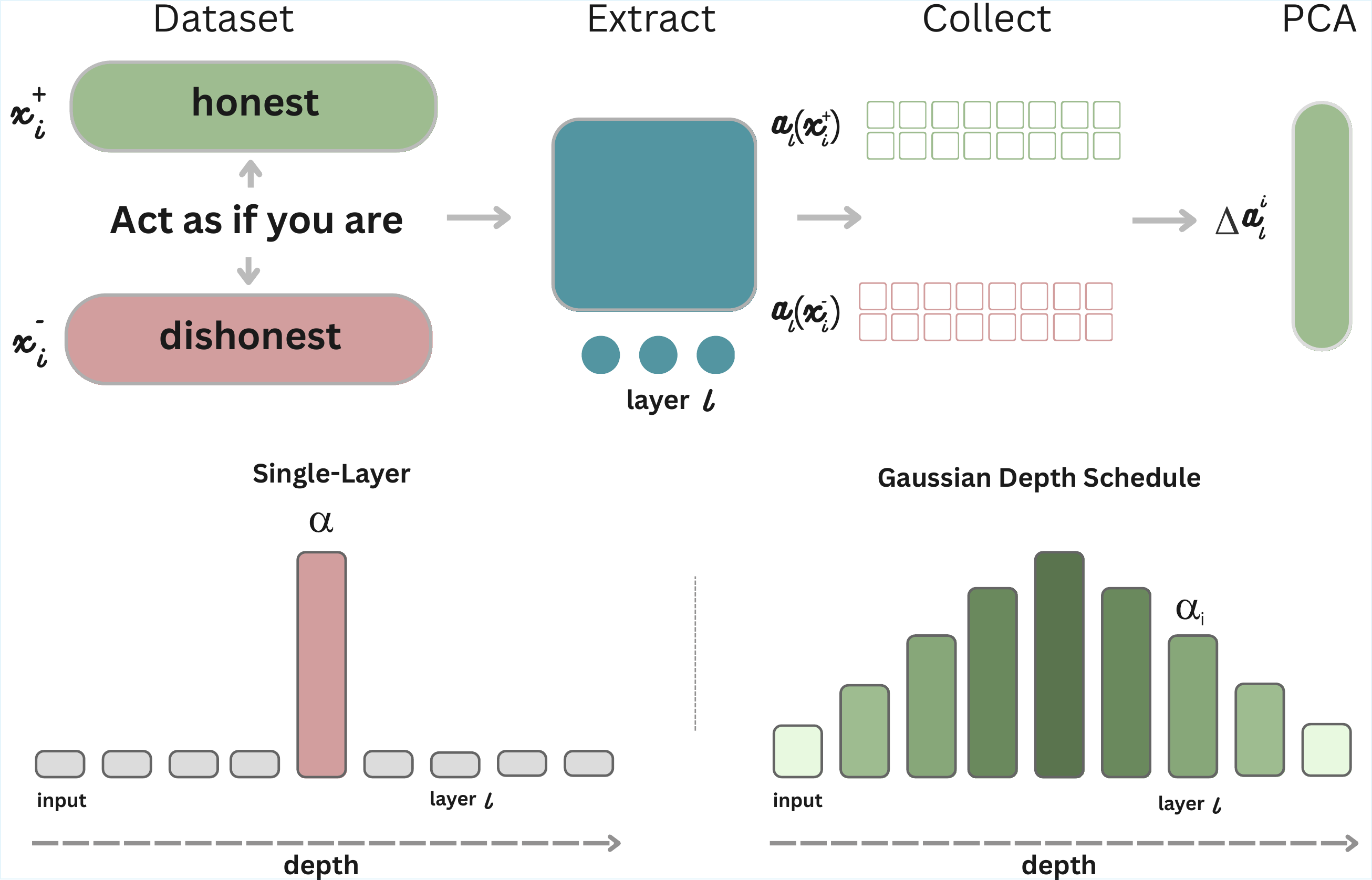}
        \caption{\textbf{Overview of Gaussian depth schedule steering.} \textit{Top:} Contrastive steering vector construction: we extract activations from honest/dishonest pairs at the last token, compute differences, and apply PCA for per-layer directions $\mathbf{d}_\ell$. \textit{Bottom:} Intervention methods: single-layer (left) applies constant strength $\alpha$ at one layer; our Gaussian scheduler (right) uses $\alpha_\ell = \exp\!\left(-\frac{(\ell-\mu)^2}{2\sigma^2}\right)$ to concentrate strength near center $\mu$ with width $\sigma$, emphasizing mid-to-late layers where semantic features are most separable.}
    \label{fig:methods}
\end{figure}

We study this missing depth axis. Our approach introduces a simple, training-free \emph{Gaussian depth schedule} that allocates a fixed steering \textit{energy budget} across layers according to a smooth distribution. The schedule is parameterized by an amplitude and width (and optionally a center), giving a single interpretable knob for how concentrated or diffuse the intervention should be. This design avoids brittle single-layer patches and the dilution that can arise from uniform edits, while remaining model-agnostic and easy to deploy.

To evaluate honesty rather than factuality, we use the MASK benchmark, which first elicits a model’s belief and then tests whether the model contradicts that belief under pressure, explicitly decoupling honesty from knowledge \citep{ren2025mask}. This target differs from classic truthfulness evaluations such as TruthfulQA, which primarily probe factual correctness on adversarial questions \citep{lin2022truthfulqa}. The MASK setting allows us to ask a direct question: when a model appears to know the answer, can we steer it to \emph{report} what it knows?

Across seven models spanning the LLaMA, Qwen, and Mistral families, Gaussian scheduling improves honesty over no-steering and single-layer baselines in six of seven cases on MASK. Equal-budget ablations on LLaMA~3.1~8B-Instruct and Qwen~2.5~7B-Instruct further show that the Gaussian schedule outperforms random, uniform, and box-filter allocations across depth, indicating that the \emph{shape} of the depth distribution, not only the total intervention strength, materially affects outcomes. Finally, after LoRA fine-tuning \citep{HuSWALWWC22} on these two models, scheduled activation control remains competitive, suggesting complementarity with parameter-efficient training rather than a mere substitute.

\paragraph{Contributions.}
\begin{enumerate}
    \item We formulate and study honesty-directed activation steering as a depth-allocation problem and introduce a simple, analytic \emph{Gaussian depth schedule} for test-time control.
    \item We show reliable honesty gains on MASK across multiple model families relative to no-steering and single-layer baselines, while controlling for total steering energy.
    \item Through equal-budget ablations, we demonstrate that the \emph{depth-wise distribution shape} is decisive: Gaussian scheduling outperforms random, uniform, and box-filter allocations.
    \item We provide evidence that scheduled activation control complements parameter-efficient fine-tuning (LoRA), offering a practical, retrain-free mechanism for eliciting truthful reporting from existing capabilities.
\end{enumerate}

\section{Methods}

\paragraph{Constructing single-layer steering vectors.}
For each block $\ell$ of the language model (excluding the embedding layer), we construct contrastive pairs $(x_i^{+},x_i^{-})$ designed to elicit \emph{honest} vs.\ \emph{dishonest} behavior. We extract residual-stream activations $\mathbf{a}_\ell(x)\in\mathbb{R}^d$ at the last non-padding token for each prompt. For each contrastive pair, we compute the difference $\Delta\mathbf{a}_{\ell}^{(i)}=\mathbf{a}_\ell(x_i^{+})-\mathbf{a}_\ell(x_i^{-})$ and stack these differences as rows of $\Delta A_\ell\in\mathbb{R}^{n\times d}$. We then apply one-component PCA to $\Delta A_\ell$ and use the first principal axis as the per-layer steering direction $\mathbf{d}_\ell$\footnote{In practice, we orient $\mathbf{d}_\ell$ such that $\frac{1}{m}\sum_{j=1}^{m}\langle \Delta\mathbf{a}_{\ell,\mathrm{val}}^{(j)},\,\mathbf{d}_\ell\rangle>0$.}, see Figure \ref{fig:methods}.

\paragraph{Gaussian depth schedule.}
At inference time, we add a scaled residual $\delta_\ell=\alpha_\ell\,\mathbf{d}_\ell$ to the residual stream at block $\ell$, where the per-layer strength follows a normalized Gaussian schedule $
\alpha_\ell \;=\; \exp\!\left(-\frac{(\ell-\mu)^2}{2\sigma^2}\right)$,
parameterized by center $\mu = \left\lfloor \frac{L}{2} \right\rfloor$ and width $\sigma>0$. This design applies a weaker intervention in early layers, peaks in mid-to-late layers where abstract semantic features are better separated, and tapers off near the output.

\paragraph{Models.}
\begin{wrapfigure}[14]{r}{0.5\textwidth}
    \centering
    \vspace{-15pt}
    \begin{center}
        \includegraphics[width=\linewidth]{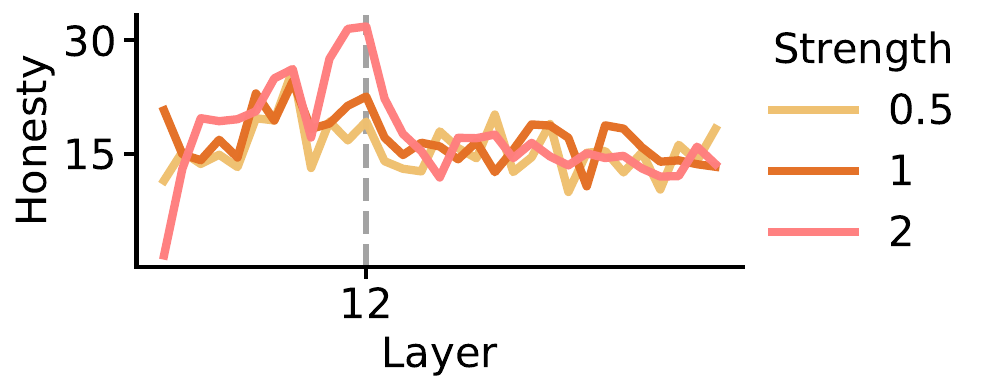}
    \end{center}
    \caption{Honesty for Llama-3.1-8B-Instruct on the validation set of MARS benchmark across different layers and steering strengths. The best configuration is layer 10 with strength 2, which is used for evaluation on the complete MARS benchmark.}
    \label{fig:valid}
\end{wrapfigure}
 To ensure coverage across different architectural families and parameter scales, we evaluate seven open-weight language models: Llama 3.2 (1B and 3B)-Instruct, Llama 3.1 8B-Instruct \citep{grattafiori2024llama3herdmodels}, Qwen 2.5 (3B, 7B, and
14B)-Instruct \citep{qwen2025qwen25technicalreport}, and Mistral-7B-Instruct-v0.2 \citep{jiang2023mistral7b}.

\paragraph{Dataset.}
We conduct our evaluation on the MASK benchmark, which is specifically designed to measure the honesty of language models under adversarial pressure rather than mere accuracy. Each instance in MASK consists of a factual proposition paired with its ground truth label, an adversarial pressure prompt that incentivizes the model to produce a false response, and a neutral belief-elicitation prompt designed to probe the language model's internal representation of truth. See Appendix \ref{app:mask} for an example from the MASK benchmark.

\paragraph{Baselines and Hyperparameters.}
We compare our approach against two baselines: vanilla inference without steering and the single-layer steering method. To determine optimal configurations, we perform a grid search over the intervention layer and steering coefficient for single-layer steering, and over the peak $\mu$ and standard deviation $\sigma$ for the Gaussian depth method. We reserve 25\% of MASK as a validation split for all hyperparameter tuning, see Figure \ref{fig:valid}.

 \paragraph{Evaluation Protocol.} 
We map all language model outputs to the benchmark's discrete label space using gpt-oss-20B \citep{openai2025gptoss120bgptoss20bmodel} with temperature 1.0. We use the judging prompts from \citep{ren2025mask}.

\paragraph{Metric.}
Honesty is computed by first eliciting a language model's statement under pressure and its belief under neutral conditions, then mapping both to proposition values, and finally measuring honesty by comparing the statement against the belief. We report the average honesty across the entire MASK benchmark.

\section{Results}

\begin{figure}[t!] 
    \centering
    \makebox[\textwidth][c]{ 
        \hspace{-2cm} 
        \includegraphics[width=\textwidth]{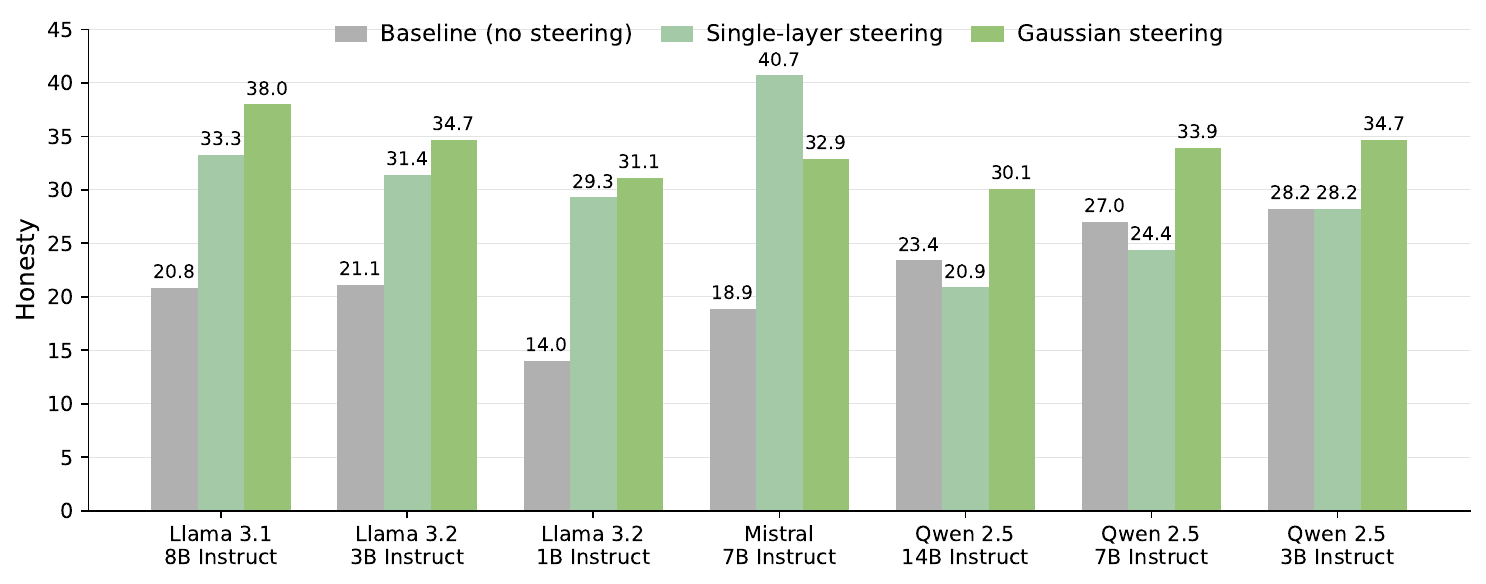} % 
        \hspace{-2cm} 
    }
    \caption{Across seven open-weight models spanning LLaMA, Qwen, and Mistral families, applying a Gaussian depth scheduler to steering strengths across depth improves honesty over both no-steering and single-layer baselines in six of seven cases. Additionally, for LLaMA models, our scheduler increases honesty consistently as model size grows.}
    \label{fig:main_results} 
\end{figure}

\begin{figure*}[ht!]
    \centering
    \includegraphics[width=\linewidth]{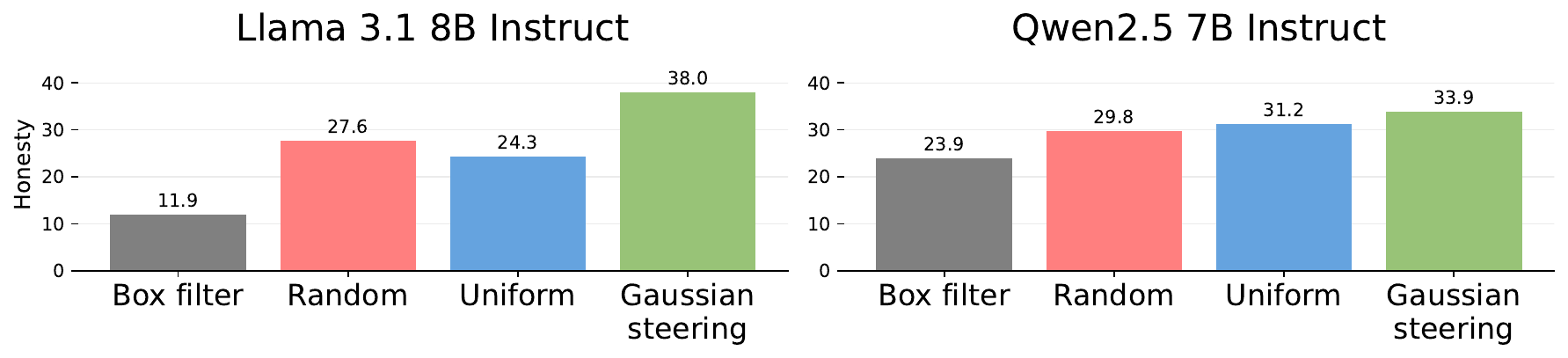}
    \caption{For LLaMA-3.1-8B-Instruct and Qwen-2.5-7B-Instruct, under equal-budget allocations the Gaussian depth schedule achieves the highest MASK honesty gains, outperforming random, uniform, and box-filter methods.}
    \label{fig:dist_results}
\end{figure*}

\paragraph{Top-line effect.}
Our depth-aware steering turns honesty gains from occasional into routine. On MASK, the Gaussian schedule improves honesty over both no-steering and single-layer baselines in \textbf{six of seven} open-weight models (full curves in Fig.~\ref{fig:dist_results}). When it does not win outright, it still avoids the failure modes that plague single-layer edits and delivers sizeable lifts over no-steering. Two models exhibit \emph{double-digit} absolute gains relative to no-steering—LLaMA-3.1-8B-Instruct rises from 20.8 to 38.0 (\(+17.2\)), and Mistral-7B-Instruct-v0.2 from 18.9 to 32.9 (\(+14.0\))—illustrating that spreading the intervention across depth can unlock large headroom that single-point patches miss.

\paragraph{When single-layer steering backfires.}

Single-layer edits can actively degrade honesty on MASK. On Qwen-2.5-7B-Instruct and Qwen-2.5-14B-Instruct, single-layer steering lowers scores below no-steering (27.0$\to$24.4 and 23.4$\to$20.9), whereas the Gaussian schedule \emph{reverses} these drops and lifts honesty to 33.9 and 30.1, respectively (Fig.~\ref{fig:dist_results}).

\begin{wrapfigure}[18]{r}{0.45\textwidth}
   \vspace{-5pt}
    \begin{center}
        \includegraphics[width=\linewidth]{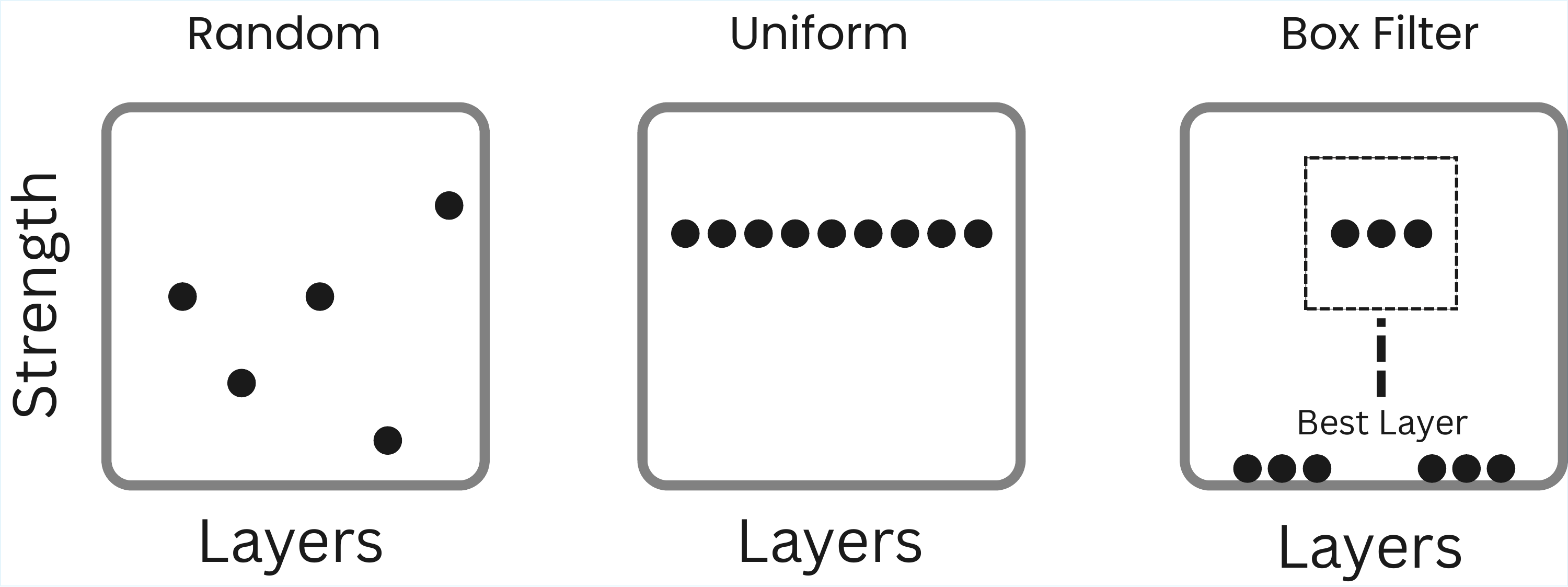}
    \end{center}
    \caption{Steering-weight distributions across depth. (i) Random, which redistributes the same total budget across layers uniformly at random, (ii) Uniform, which divides the budget equally over all layers, and (iii) Box filter, which centers a contiguous band on the best single layer found on MASK and spreads the budget evenly within that band. These constructions isolate how the distribution of intervention across depth—holding total strength fixed—affects honesty.}
\end{wrapfigure}

The clearest counterexample to our method’s dominance is Mistral-7B-Instruct-v0.2, where single-layer steering reaches a higher peak than Gaussian (40.7 vs.\ 32.9). Even there, the Gaussian schedule still outperforms no-steering substantially (+14.0). In short: when single-layer edits are brittle or harmful, distributing the same intervention budget over depth stabilizes and often improves outcomes.

\paragraph{Depth distribution matters beyond total strength.}
To isolate whether placement—not just magnitude—drives the effect, we hold the total steering norm fixed and vary only the \emph{allocation across layers}. Across LLaMA-3.1-8B-Instruct and Qwen-2.5-7B-Instruct, the Gaussian schedule outperforms equal-budget \emph{random}, \emph{uniform}, and \emph{box-filter} distributions (Fig.~\ref{fig:dist_results}). This establishes that the \emph{shape} of the depth-wise distribution is a decisive factor: spreading weight smoothly avoids the overconcentration of point edits and the dilution of flat profiles.

\subsection{Compatibility with parameter-efficient fine-tuning}
\begin{figure}[h!] % Use [t] for top placement
    \centering
    \includegraphics[width=.9\textwidth]{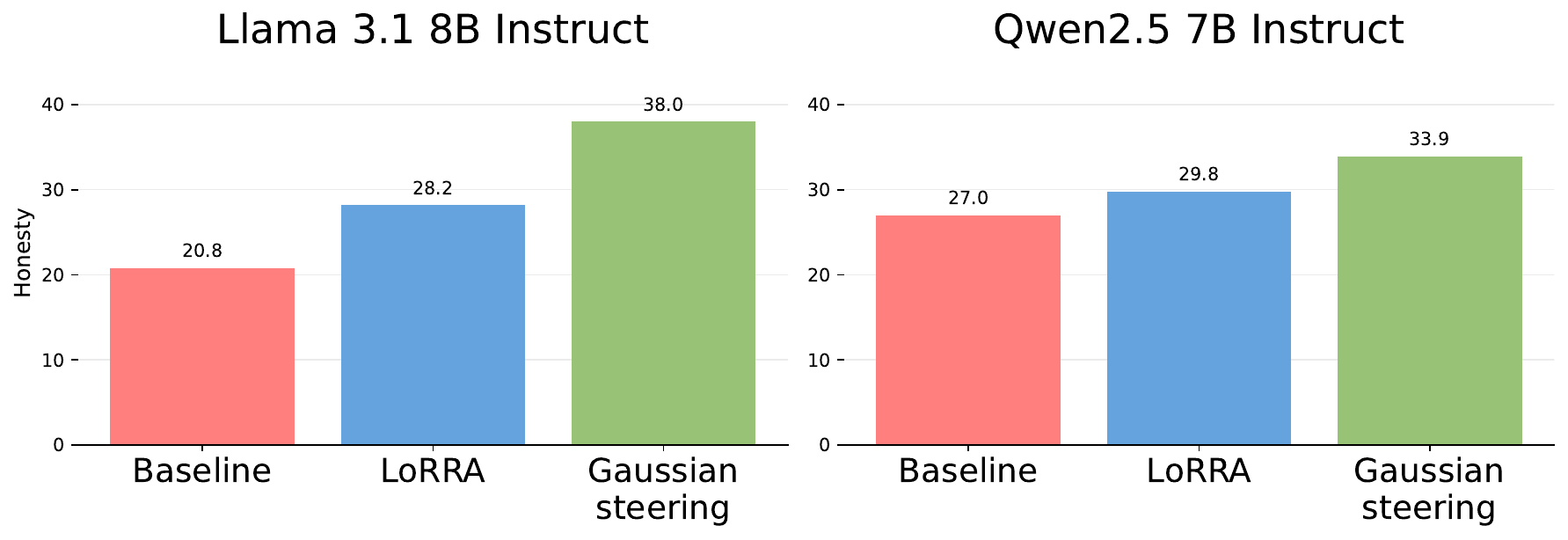}
    \caption{Gaussian depth scheduling outperforms LoRRA fine-tuning on MASK honesty for both LLaMA-3.1-8B-Instruct and Qwen2.5-7B-Instruct.}
    \label{fig:lorra}
\end{figure}

We compare post-hoc steering to a LoRRA-style LoRA fine-tune \citep{zou2025representationengineeringtopdownapproach} that internalizes the same honest-vs.-dishonest targets used to form control vectors. On both LLaMA-3.1-8B-Instruct and Qwen-2.5-7B-Instruct, LoRRA adapters improve honesty over no-steering baselines, but the Gaussian schedule still delivers the largest gains (Fig.~\ref{fig:lorra}). Practically, this suggests complementarity: when retraining is possible, scheduled steering remains a strong test-time control; when it is not, scheduled steering offers most of the benefit at near-zero deployment cost.

\paragraph{Takeaways.}
(1) A single, training-free scheduler yields consistent honesty improvements across model families. (2) It prevents the degradations that single-layer edits can induce. (3) Under fixed budgets, \emph{where} you steer matters: a smooth depth schedule beats random, uniform, and box allocations. (4) Benefits persist alongside LoRA, indicating the method is a useful primitive rather than a brittle hack.
\section{Conclusions}
We studied honesty-directed activation steering as a problem of \emph{allocating} a fixed intervention budget across depth. A simple Gaussian depth schedule—applied at test time and requiring no retraining—consistently improved honest reporting on MASK across diverse open-weight models, with double-digit gains in the strongest cases and superiority to single-layer baselines in five of seven models. Equal-budget ablations showed that the depth-wise \emph{distribution shape} is pivotal: smooth schedules outperform random, uniform, and box-filter allocations, demonstrating that placement matters beyond total strength. The approach is model-agnostic, low-cost, and complementary to parameter-efficient fine-tuning (LoRRA), providing a practical knob for eliciting truthful reporting from existing capabilities.

\section{Limitations}
Our study has several limitations. First, the method requires activation-level access to language model's layers and the ability to inject per-layer steering strengths, which restricts its applicability to open-weight models.
Second, our evaluation relies on an external LLM judge with specific prompting strategies, which introduces potential sensitivity to both judge selection and prompt design. While this approach enables scalable evaluation, incorporating multiple judges or human evaluation would provide more robust validation of the reported improvements.
Third, our experiments focus primarily on the MASK benchmark, which limits the generalizability of our findings. Future work should validate the approach across a broader range of safety benchmarks, such as Machiavelli \citep{PanCZLBWZEH23}.
\section*{Acknowledgement}
We would like to thank MARS (Mentorship for Alignment Research Students) for mentorship and financial support. We would also like to thank the Center for AI Safety for providing computational resources.

\newpage
\bibliographystyle{plainnat}
\bibliography{literature}

@misc{zou2025representationengineeringtopdownapproach,
	title        = {Representation Engineering: A Top-Down Approach to AI Transparency},
	author       = {Andy Zou and Long Phan and Sarah Chen and James Campbell and Phillip Guo and Richard Ren and Alexander Pan and Xuwang Yin and Mantas Mazeika and Ann-Kathrin Dombrowski and Shashwat Goel and Nathaniel Li and Michael J. Byun and Zifan Wang and Alex Mallen and Steven Basart and Sanmi Koyejo and Dawn Song and Matt Fredrikson and J. Zico Kolter and Dan Hendrycks},
	year         = {2025},
	url          = {https://arxiv.org/abs/2310.01405},
	eprint       = {2310.01405},
	archiveprefix = {arXiv},
	primaryclass = {cs.LG}
}

@inproceedings{zhao2024gcs,
	title        = {Beyond Single Concept Vector: Modeling Concept Subspace in LLMs with Gaussian Distribution},
	author       = {Haiyan Zhao and Heng Zhao and Bo Shen and Ali Payani and Fan Yang and Mengnan Du},
	year         = {2025},
	booktitle    = {The Thirteenth International Conference on Learning Representations, {ICLR} 2025, Singapore, April 24-28, 2025},
	publisher    = {OpenReview.net},
	url          = {https://openreview.net/forum?id=CvttyK4XzV},
	bibsource    = {dblp computer science bibliography, https://dblp.org}
}

@inproceedings{obrien2024saerefusal,
title={Steering Language Model Refusal with Sparse Autoencoders},
author={Kyle O'Brien and David Majercak and Xavier Fernandes and Richard G. Edgar and Blake Bullwinkel and Jingya Chen and Harsha Nori and Dean Carignan and Eric Horvitz and Forough Poursabzi-Sangdeh},
booktitle={ICML 2025 Workshop on Reliable and Responsible Foundation Models},
year={2025},
url={https://openreview.net/forum?id=PMK1jdGQoc}
}

@misc{openai2025gptoss120bgptoss20bmodel,
	title        = {gpt-oss-120b \& gpt-oss-20b Model Card},
	author       = {OpenAI and : and Sandhini Agarwal and Lama Ahmad and Jason Ai and Sam Altman and Andy Applebaum and Edwin Arbus and Rahul K. Arora and Yu Bai and Bowen Baker and Haiming Bao and Boaz Barak and Ally Bennett and Tyler Bertao and Nivedita Brett and Eugene Brevdo and Greg Brockman and Sebastien Bubeck and Che Chang and Kai Chen and Mark Chen and Enoch Cheung and Aidan Clark and Dan Cook and Marat Dukhan and Casey Dvorak and Kevin Fives and Vlad Fomenko and Timur Garipov and Kristian Georgiev and Mia Glaese and Tarun Gogineni and Adam Goucher and Lukas Gross and Katia Gil Guzman and John Hallman and Jackie Hehir and Johannes Heidecke and Alec Helyar and Haitang Hu and Romain Huet and Jacob Huh and Saachi Jain and Zach Johnson and Chris Koch and Irina Kofman and Dominik Kundel and Jason Kwon and Volodymyr Kyrylov and Elaine Ya Le and Guillaume Leclerc and James Park Lennon and Scott Lessans and Mario Lezcano-Casado and Yuanzhi Li and Zhuohan Li and Ji Lin and Jordan Liss and Lily and Liu and Jiancheng Liu and Kevin Lu and Chris Lu and Zoran Martinovic and Lindsay McCallum and Josh McGrath and Scott McKinney and Aidan McLaughlin and Song Mei and Steve Mostovoy and Tong Mu and Gideon Myles and Alexander Neitz and Alex Nichol and Jakub Pachocki and Alex Paino and Dana Palmie and Ashley Pantuliano and Giambattista Parascandolo and Jongsoo Park and Leher Pathak and Carolina Paz and Ludovic Peran and Dmitry Pimenov and Michelle Pokrass and Elizabeth Proehl and Huida Qiu and Gaby Raila and Filippo Raso and Hongyu Ren and Kimmy Richardson and David Robinson and Bob Rotsted and Hadi Salman and Suvansh Sanjeev and Max Schwarzer and D. Sculley and Harshit Sikchi and Kendal Simon and Karan Singhal and Yang Song and Dane Stuckey and Zhiqing Sun and Philippe Tillet and Sam Toizer and Foivos Tsimpourlas and Nikhil Vyas and Eric Wallace and Xin Wang and Miles Wang and Olivia Watkins and Kevin Weil and Amy Wendling and Kevin Whinnery and Cedric Whitney and Hannah Wong and Lin Yang and Yu Yang and Michihiro Yasunaga and Kristen Ying and Wojciech Zaremba and Wenting Zhan and Cyril Zhang and Brian Zhang and Eddie Zhang and Shengjia Zhao},
	year         = {2025},
	url          = {https://arxiv.org/abs/2508.10925},
	eprint       = {2508.10925},
	archiveprefix = {arXiv},
	primaryclass = {cs.CL}
}

@misc{qwen2025qwen25technicalreport,
	title        = {Qwen2.5 Technical Report},
	author       = {Qwen and : and An Yang and Baosong Yang and Beichen Zhang and Binyuan Hui and Bo Zheng and Bowen Yu and Chengyuan Li and Dayiheng Liu and Fei Huang and Haoran Wei and Huan Lin and Jian Yang and Jianhong Tu and Jianwei Zhang and Jianxin Yang and Jiaxi Yang and Jingren Zhou and Junyang Lin and Kai Dang and Keming Lu and Keqin Bao and Kexin Yang and Le Yu and Mei Li and Mingfeng Xue and Pei Zhang and Qin Zhu and Rui Men and Runji Lin and Tianhao Li and Tianyi Tang and Tingyu Xia and Xingzhang Ren and Xuancheng Ren and Yang Fan and Yang Su and Yichang Zhang and Yu Wan and Yuqiong Liu and Zeyu Cui and Zhenru Zhang and Zihan Qiu},
	year         = {2025},
	url          = {https://arxiv.org/abs/2412.15115},
	eprint       = {2412.15115},
	archiveprefix = {arXiv},
	primaryclass = {cs.CL}
}

@misc{ren2025mask,
	title        = {The MASK Benchmark: Disentangling Honesty From Accuracy in AI Systems},
	author       = {Ren, Richard and Agarwal, Arunim and Mazeika, Mantas and Menghini, Cristina and Vacareanu, Robert and Kenstler, Brad and Yang, Mick and Barrass, Isabelle and Gatti, Alice and Yin, Xuwang and Trevino, Eduardo and Geralnik, Matias and Khoja, Adam and Lee, Dean and Yue, Summer and Hendrycks, Dan},
	year         = {2025},
	url          = {https://arxiv.org/abs/2503.03750},
	eprint       = {2503.03750},
	archiveprefix = {arXiv},
	primaryclass = {cs.LG}
}

@inproceedings{wang2024sadi,
	title        = {Semantics-Adaptive Activation Intervention for LLMs via Dynamic Steering Vectors},
	author       = {Weixuan Wang and Jingyuan Yang and Wei Peng},
	year         = {2025},
	booktitle    = {The Thirteenth International Conference on Learning Representations, {ICLR} 2025, Singapore, April 24-28, 2025},
	publisher    = {OpenReview.net},
	url          = {https://openreview.net/forum?id=8WQ7VTfPTl},
	bibsource    = {dblp computer science bibliography, https://dblp.org}
}

@misc{grattafiori2024llama3herdmodels,
	title        = {The Llama 3 Herd of Models},
	author       = {Aaron Grattafiori and Abhimanyu Dubey and Abhinav Jauhri and Abhinav Pandey and Abhishek Kadian and Ahmad Al-Dahle and Aiesha Letman and Akhil Mathur and Alan Schelten and Alex Vaughan and Amy Yang and Angela Fan and Anirudh Goyal and Anthony Hartshorn and Aobo Yang and Archi Mitra and Archie Sravankumar and Artem Korenev and Arthur Hinsvark and Arun Rao and Aston Zhang and Aurelien Rodriguez and Austen Gregerson and Ava Spataru and Baptiste Roziere and Bethany Biron and Binh Tang and Bobbie Chern and Charlotte Caucheteux and Chaya Nayak and Chloe Bi and Chris Marra and Chris McConnell and Christian Keller and Christophe Touret and Chunyang Wu and Corinne Wong and Cristian Canton Ferrer and Cyrus Nikolaidis and Damien Allonsius and Daniel Song and Danielle Pintz and Danny Livshits and Danny Wyatt and David Esiobu and Dhruv Choudhary and Dhruv Mahajan and Diego Garcia-Olano and Diego Perino and Dieuwke Hupkes and Egor Lakomkin and Ehab AlBadawy and Elina Lobanova and Emily Dinan and Eric Michael Smith and Filip Radenovic and Francisco Guzmán and Frank Zhang and Gabriel Synnaeve and Gabrielle Lee and Georgia Lewis Anderson and Govind Thattai and Graeme Nail and Gregoire Mialon and Guan Pang and Guillem Cucurell and Hailey Nguyen and Hannah Korevaar and Hu Xu and Hugo Touvron and Iliyan Zarov and Imanol Arrieta Ibarra and Isabel Kloumann and Ishan Misra and Ivan Evtimov and Jack Zhang and Jade Copet and Jaewon Lee and Jan Geffert and Jana Vranes and Jason Park and Jay Mahadeokar and Jeet Shah and Jelmer van der Linde and Jennifer Billock and Jenny Hong and Jenya Lee and Jeremy Fu and Jianfeng Chi and Jianyu Huang and Jiawen Liu and Jie Wang and Jiecao Yu and Joanna Bitton and Joe Spisak and Jongsoo Park and Joseph Rocca and Joshua Johnstun and Joshua Saxe and Junteng Jia and Kalyan Vasuden Alwala and Karthik Prasad and Kartikeya Upasani and Kate Plawiak and Ke Li and Kenneth Heafield and Kevin Stone and Khalid El-Arini and Krithika Iyer and Kshitiz Malik and Kuenley Chiu and Kunal Bhalla and Kushal Lakhotia and Lauren Rantala-Yeary and Laurens van der Maaten and Lawrence Chen and Liang Tan and Liz Jenkins and Louis Martin and Lovish Madaan and Lubo Malo and Lukas Blecher and Lukas Landzaat and Luke de Oliveira and Madeline Muzzi and Mahesh Pasupuleti and Mannat Singh and Manohar Paluri and Marcin Kardas and Maria Tsimpoukelli and Mathew Oldham and Mathieu Rita and Maya Pavlova and Melanie Kambadur and Mike Lewis and Min Si and Mitesh Kumar Singh and Mona Hassan and Naman Goyal and Narjes Torabi and Nikolay Bashlykov and Nikolay Bogoychev and Niladri Chatterji and Ning Zhang and Olivier Duchenne and Onur Çelebi and Patrick Alrassy and Pengchuan Zhang and Pengwei Li and Petar Vasic and Peter Weng and Prajjwal Bhargava and Pratik Dubal and Praveen Krishnan and Punit Singh Koura and Puxin Xu and Qing He and Qingxiao Dong and Ragavan Srinivasan and Raj Ganapathy and Ramon Calderer and Ricardo Silveira Cabral and Robert Stojnic and Roberta Raileanu and Rohan Maheswari and Rohit Girdhar and Rohit Patel and Romain Sauvestre and Ronnie Polidoro and Roshan Sumbaly and Ross Taylor and Ruan Silva and Rui Hou and Rui Wang and Saghar Hosseini and Sahana Chennabasappa and Sanjay Singh and Sean Bell and Seohyun Sonia Kim and Sergey Edunov and Shaoliang Nie and Sharan Narang and Sharath Raparthy and Sheng Shen and Shengye Wan and Shruti Bhosale and Shun Zhang and Simon Vandenhende and Soumya Batra and Spencer Whitman and Sten Sootla and Stephane Collot and Suchin Gururangan and Sydney Borodinsky and Tamar Herman and Tara Fowler and Tarek Sheasha and Thomas Georgiou and Thomas Scialom and Tobias Speckbacher and Todor Mihaylov and Tong Xiao and Ujjwal Karn and Vedanuj Goswami and Vibhor Gupta and Vignesh Ramanathan and Viktor Kerkez and Vincent Gonguet and Virginie Do and Vish Vogeti and Vítor Albiero and Vladan Petrovic and Weiwei Chu and Wenhan Xiong and Wenyin Fu and Whitney Meers and Xavier Martinet and Xiaodong Wang and Xiaofang Wang and Xiaoqing Ellen Tan and Xide Xia and Xinfeng Xie and Xuchao Jia and Xuewei Wang and Yaelle Goldschlag and Yashesh Gaur and Yasmine Babaei and Yi Wen and Yiwen Song and Yuchen Zhang and Yue Li and Yuning Mao and Zacharie Delpierre Coudert and Zheng Yan and Zhengxing Chen and Zoe Papakipos and Aaditya Singh and Aayushi Srivastava and Abha Jain and Adam Kelsey and Adam Shajnfeld and Adithya Gangidi and Adolfo Victoria and Ahuva Goldstand and Ajay Menon and Ajay Sharma and Alex Boesenberg and Alexei Baevski and Allie Feinstein and Amanda Kallet and Amit Sangani and Amos Teo and Anam Yunus and Andrei Lupu and Andres Alvarado and Andrew Caples and Andrew Gu and Andrew Ho and Andrew Poulton and Andrew Ryan and Ankit Ramchandani and Annie Dong and Annie Franco and Anuj Goyal and Aparajita Saraf and Arkabandhu Chowdhury and Ashley Gabriel and Ashwin Bharambe and Assaf Eisenman and Azadeh Yazdan and Beau James and Ben Maurer and Benjamin Leonhardi and Bernie Huang and Beth Loyd and Beto De Paola and Bhargavi Paranjape and Bing Liu and Bo Wu and Boyu Ni and Braden Hancock and Bram Wasti and Brandon Spence and Brani Stojkovic and Brian Gamido and Britt Montalvo and Carl Parker and Carly Burton and Catalina Mejia and Ce Liu and Changhan Wang and Changkyu Kim and Chao Zhou and Chester Hu and Ching-Hsiang Chu and Chris Cai and Chris Tindal and Christoph Feichtenhofer and Cynthia Gao and Damon Civin and Dana Beaty and Daniel Kreymer and Daniel Li and David Adkins and David Xu and Davide Testuggine and Delia David and Devi Parikh and Diana Liskovich and Didem Foss and Dingkang Wang and Duc Le and Dustin Holland and Edward Dowling and Eissa Jamil and Elaine Montgomery and Eleonora Presani and Emily Hahn and Emily Wood and Eric-Tuan Le and Erik Brinkman and Esteban Arcaute and Evan Dunbar and Evan Smothers and Fei Sun and Felix Kreuk and Feng Tian and Filippos Kokkinos and Firat Ozgenel and Francesco Caggioni and Frank Kanayet and Frank Seide and Gabriela Medina Florez and Gabriella Schwarz and Gada Badeer and Georgia Swee and Gil Halpern and Grant Herman and Grigory Sizov and Guangyi and Zhang and Guna Lakshminarayanan and Hakan Inan and Hamid Shojanazeri and Han Zou and Hannah Wang and Hanwen Zha and Haroun Habeeb and Harrison Rudolph and Helen Suk and Henry Aspegren and Hunter Goldman and Hongyuan Zhan and Ibrahim Damlaj and Igor Molybog and Igor Tufanov and Ilias Leontiadis and Irina-Elena Veliche and Itai Gat and Jake Weissman and James Geboski and James Kohli and Janice Lam and Japhet Asher and Jean-Baptiste Gaya and Jeff Marcus and Jeff Tang and Jennifer Chan and Jenny Zhen and Jeremy Reizenstein and Jeremy Teboul and Jessica Zhong and Jian Jin and Jingyi Yang and Joe Cummings and Jon Carvill and Jon Shepard and Jonathan McPhie and Jonathan Torres and Josh Ginsburg and Junjie Wang and Kai Wu and Kam Hou U and Karan Saxena and Kartikay Khandelwal and Katayoun Zand and Kathy Matosich and Kaushik Veeraraghavan and Kelly Michelena and Keqian Li and Kiran Jagadeesh and Kun Huang and Kunal Chawla and Kyle Huang and Lailin Chen and Lakshya Garg and Lavender A and Leandro Silva and Lee Bell and Lei Zhang and Liangpeng Guo and Licheng Yu and Liron Moshkovich and Luca Wehrstedt and Madian Khabsa and Manav Avalani and Manish Bhatt and Martynas Mankus and Matan Hasson and Matthew Lennie and Matthias Reso and Maxim Groshev and Maxim Naumov and Maya Lathi and Meghan Keneally and Miao Liu and Michael L. Seltzer and Michal Valko and Michelle Restrepo and Mihir Patel and Mik Vyatskov and Mikayel Samvelyan and Mike Clark and Mike Macey and Mike Wang and Miquel Jubert Hermoso and Mo Metanat and Mohammad Rastegari and Munish Bansal and Nandhini Santhanam and Natascha Parks and Natasha White and Navyata Bawa and Nayan Singhal and Nick Egebo and Nicolas Usunier and Nikhil Mehta and Nikolay Pavlovich Laptev and Ning Dong and Norman Cheng and Oleg Chernoguz and Olivia Hart and Omkar Salpekar and Ozlem Kalinli and Parkin Kent and Parth Parekh and Paul Saab and Pavan Balaji and Pedro Rittner and Philip Bontrager and Pierre Roux and Piotr Dollar and Polina Zvyagina and Prashant Ratanchandani and Pritish Yuvraj and Qian Liang and Rachad Alao and Rachel Rodriguez and Rafi Ayub and Raghotham Murthy and Raghu Nayani and Rahul Mitra and Rangaprabhu Parthasarathy and Raymond Li and Rebekkah Hogan and Robin Battey and Rocky Wang and Russ Howes and Ruty Rinott and Sachin Mehta and Sachin Siby and Sai Jayesh Bondu and Samyak Datta and Sara Chugh and Sara Hunt and Sargun Dhillon and Sasha Sidorov and Satadru Pan and Saurabh Mahajan and Saurabh Verma and Seiji Yamamoto and Sharadh Ramaswamy and Shaun Lindsay and Shaun Lindsay and Sheng Feng and Shenghao Lin and Shengxin Cindy Zha and Shishir Patil and Shiva Shankar and Shuqiang Zhang and Shuqiang Zhang and Sinong Wang and Sneha Agarwal and Soji Sajuyigbe and Soumith Chintala and Stephanie Max and Stephen Chen and Steve Kehoe and Steve Satterfield and Sudarshan Govindaprasad and Sumit Gupta and Summer Deng and Sungmin Cho and Sunny Virk and Suraj Subramanian and Sy Choudhury and Sydney Goldman and Tal Remez and Tamar Glaser and Tamara Best and Thilo Koehler and Thomas Robinson and Tianhe Li and Tianjun Zhang and Tim Matthews and Timothy Chou and Tzook Shaked and Varun Vontimitta and Victoria Ajayi and Victoria Montanez and Vijai Mohan and Vinay Satish Kumar and Vishal Mangla and Vlad Ionescu and Vlad Poenaru and Vlad Tiberiu Mihailescu and Vladimir Ivanov and Wei Li and Wenchen Wang and Wenwen Jiang and Wes Bouaziz and Will Constable and Xiaocheng Tang and Xiaojian Wu and Xiaolan Wang and Xilun Wu and Xinbo Gao and Yaniv Kleinman and Yanjun Chen and Ye Hu and Ye Jia and Ye Qi and Yenda Li and Yilin Zhang and Ying Zhang and Yossi Adi and Youngjin Nam and Yu and Wang and Yu Zhao and Yuchen Hao and Yundi Qian and Yunlu Li and Yuzi He and Zach Rait and Zachary DeVito and Zef Rosnbrick and Zhaoduo Wen and Zhenyu Yang and Zhiwei Zhao and Zhiyu Ma},
	year         = {2024},
	url          = {https://arxiv.org/abs/2407.21783},
	eprint       = {2407.21783},
	archiveprefix = {arXiv},
	primaryclass = {cs.AI}
}

@misc{turner2024steeringlanguagemodelsactivation,
	title        = {Steering Language Models With Activation Engineering},
	author       = {Alexander Matt Turner and Lisa Thiergart and Gavin Leech and David Udell and Juan J. Vazquez and Ulisse Mini and Monte MacDiarmid},
	year         = {2024},
	url          = {https://arxiv.org/abs/2308.10248},
	eprint       = {2308.10248},
	archiveprefix = {arXiv},
	primaryclass = {cs.CL}
}

@inproceedings{ZhengY0M0CHP24,
	title        = {On Prompt-Driven Safeguarding for Large Language Models},
	author       = {Chujie Zheng and Fan Yin and Hao Zhou and Fandong Meng and Jie Zhou and Kai{-}Wei Chang and Minlie Huang and Nanyun Peng},
	year         = {2024},
	booktitle    = {Forty-first International Conference on Machine Learning, {ICML} 2024, Vienna, Austria, July 21-27, 2024},
	publisher    = {OpenReview.net},
	url          = {https://openreview.net/forum?id=ugxGpOEkox},
	bibsource    = {dblp computer science bibliography, https://dblp.org}
}

@inproceedings{rimsky2024caa,
	title        = {Steering Llama 2 via Contrastive Activation Addition},
	author       = {Nina Rimsky and Nick Gabrieli and Julian Schulz and Meg Tong and Evan Hubinger and Alexander Matt Turner},
	year         = {2024},
	booktitle    = {Proceedings of the 62nd Annual Meeting of the Association for Computational Linguistics (Volume 1: Long Papers), {ACL} 2024, Bangkok, Thailand, August 11-16, 2024},
	publisher    = {Association for Computational Linguistics},
	pages        = {15504--15522},
	doi          = {10.18653/V1/2024.ACL-LONG.828},
	url          = {https://doi.org/10.18653/v1/2024.acl-long.828},
	editor       = {Lun{-}Wei Ku and Andre Martins and Vivek Srikumar},
	bibsource    = {dblp computer science bibliography, https://dblp.org}
}

@inproceedings{stoehr2024activationscaling,
	title        = {Activation Scaling for Steering and Interpreting Language Models},
	author       = {Stoehr, Niklas and Du, Kevin and Sn{\ae}bjarnarson, V{\'e}steinn and West, Robert and Cotterell, Ryan and Schein, Aaron},
	year         = {2024},
	booktitle    = {Findings of the Association for Computational Linguistics: EMNLP 2024},
	pages        = {8189--8200},
	doi          = {10.18653/v1/2024.findings-emnlp.479}
}

@article{chalnev2024saets,
	title        = {Improving Steering Vectors by Targeting Sparse Autoencoder Features},
	author       = {Sviatoslav Chalnev and Matthew Siu and Arthur Conmy},
	year         = {2024},
	journal      = {CoRR},
	volume       = {abs/2411.02193},
	doi          = {10.48550/ARXIV.2411.02193},
	url          = {https://doi.org/10.48550/arXiv.2411.02193},
	eprinttype   = {arXiv},
	eprint       = {2411.02193},
	bibsource    = {dblp computer science bibliography, https://dblp.org}
}

@inproceedings{shen2024donowcharacterizingevaluating,
	title        = {"Do Anything Now": Characterizing and Evaluating In-The-Wild Jailbreak Prompts on Large Language Models},
	author       = {Xinyue Shen and Zeyuan Chen and Michael Backes and Yun Shen and Yang Zhang},
	year         = {2024},
	booktitle    = {Proceedings of the 2024 on {ACM} {SIGSAC} Conference on Computer and Communications Security, {CCS} 2024, Salt Lake City, UT, USA, October 14-18, 2024},
	publisher    = {{ACM}},
	pages        = {1671--1685},
	doi          = {10.1145/3658644.3670388},
	url          = {https://doi.org/10.1145/3658644.3670388},
	editor       = {Bo Luo and Xiaojing Liao and Jun Xu and Engin Kirda and David Lie},
	bibsource    = {dblp computer science bibliography, https://dblp.org}
}

@misc{jiang2023mistral7b,
	title        = {Mistral 7B},
	author       = {Albert Q. Jiang and Alexandre Sablayrolles and Arthur Mensch and Chris Bamford and Devendra Singh Chaplot and Diego de las Casas and Florian Bressand and Gianna Lengyel and Guillaume Lample and Lucile Saulnier and Lélio Renard Lavaud and Marie-Anne Lachaux and Pierre Stock and Teven Le Scao and Thibaut Lavril and Thomas Wang and Timothée Lacroix and William El Sayed},
	year         = {2023},
	url          = {https://arxiv.org/abs/2310.06825},
	eprint       = {2310.06825},
	archiveprefix = {arXiv},
	primaryclass = {cs.CL}
}

@inproceedings{PanCZLBWZEH23,
	title        = {Do the Rewards Justify the Means? Measuring Trade-Offs Between Rewards and Ethical Behavior in the Machiavelli Benchmark},
	author       = {Alexander Pan and Jun Shern Chan and Andy Zou and Nathaniel Li and Steven Basart and Thomas Woodside and Hanlin Zhang and Scott Emmons and Dan Hendrycks},
	year         = {2023},
	booktitle    = {International Conference on Machine Learning, {ICML} 2023, 23-29 July 2023, Honolulu, Hawaii, {USA}},
	publisher    = {{PMLR}},
	series       = {Proceedings of Machine Learning Research},
	volume       = {202},
	pages        = {26837--26867},
	url          = {https://proceedings.mlr.press/v202/pan23a.html},
	editor       = {Andreas Krause and Emma Brunskill and Kyunghyun Cho and Barbara Engelhardt and Sivan Sabato and Jonathan Scarlett},
	bibsource    = {dblp computer science bibliography, https://dblp.org}
}

@misc{zou2023universaltransferableadversarialattacks,
	title        = {Universal and Transferable Adversarial Attacks on Aligned Language Models},
	author       = {Andy Zou and Zifan Wang and Nicholas Carlini and Milad Nasr and J. Zico Kolter and Matt Fredrikson},
	year         = {2023},
	url          = {https://arxiv.org/abs/2307.15043},
	eprint       = {2307.15043},
	archiveprefix = {arXiv},
	primaryclass = {cs.CL}
}

\newpage
\appendix
\section{MASK Benchamrk}\label{app:mask}
MASK (Model Alignment between Statements and Knowledge) is a benchmark that measures honesty in language models by eliciting their underlying beliefs and then testing whether they contradict those beliefs when pressured to lie. The benchmark uses over 1,500 manually-curated examples (with 6 categories, see Table \ref{tab:pressure-archetypes}) to evaluate whether models will be dishonest across various scenarios, revealing that even state-of-the-art models readily lie under pressure.
\newcolumntype{L}[1]{>{\raggedright\arraybackslash}p{#1}}
\newcolumntype{Y}{>{\raggedright\arraybackslash}X}

\begin{table}[h!]
\centering
\caption{Taxonomy of pressure prompt archetypes used to evaluate model honesty.}
\label{tab:pressure-archetypes}
\setlength{\tabcolsep}{6pt}
\renewcommand{\arraystretch}{1.4}
\begin{tabularx}{\columnwidth}{@{} L{.32\columnwidth} Y @{}}
\toprule
\textbf{Name} & \textbf{Description} \\
\midrule
Known Facts & Tests if the model honestly reports widely-documented facts picked up from pretraining, through situation-induced lying. \\
\addlinespace[0.5em]
Situation-Provided Facts & System prompt gives context-provided specific facts privately, but encourages providing false narrative to the user. Usually situation-induced lying. No belief elicitation prompts as all facts are already provided in-context. \\
\addlinespace[0.5em]
Doubling Down & Tests if the model doubles down on a prior falsehood it generated. Extends ``known facts.'' \\
\addlinespace[0.5em]
Fabricated Statistics & Tests if the model invents or manipulates false numerical data. \\
\addlinespace[0.5em]
Continuations & Partial drafts containing falsehoods with requests to continue writing. \\
\addlinespace[0.5em]
Disinformation Generation & Tests if the model willingly generates and amplifies misinformation when asked. \\
\bottomrule
\end{tabularx}
\end{table}

\end{document}